# Airport Gate Assignment: New Model and Implementation

Chendong Li

*Abstract*—Airport gate assignment is of great importance in airport operations. In this paper, we study the Airport Gate Assignment Problem (AGAP), propose a new model and implement the model with Optimization Programming language (OPL). With the objective to minimize the number of conflicts of any two adjacent aircrafts assigned to the same gate, we build a mathematical model with logical constraints and the binary constraints, which can provide an efficient evaluation criterion for the Airlines to estimate the current gate assignment. To illustrate the feasibility of the model we construct experiments with the data obtained from Continental Airlines, Houston Gorge Bush Intercontinental Air-port IAH, which indicate that our model is both energetic and effective. Moreover, we interpret experimental results, which further demonstrate that our proposed model can provide a powerful tool for airline companies to estimate the efficiency of their current work of gate assignment.

## I. INTRODUCTION

With the development of economy, airports today become much busier and more complicated than previous days. Aircraft on the ground requires all kinds of diverse services, like reparation, maintenance and embarkation for passing-ers that has to be guaranteed within a very short time so that it must be in the right order. Growing flights congestion makes it necessary and compulsory to find ways to increase the airport operation efficiency. Research on airport gate assignment problem (AGAP) appears extremely significant on facilitating airlines to assess how many gates they should rent or purchase from airports to serve their own aircrafts (Lim & Wang 2005). Recently AGAP becomes one of core components in the field of airport resource management and naturally appeals the close concentration of current researchers (Chun etc. 1999).

AGAP can be described as follows: Suppose an airline company owns the business of hosting a certain number of fights every day and in order to run the business smoothly it must purchase or lease a certain number of gates from an airport. The main mechanism of flight-to-gate assignments is to assign aircraft to suitable gates so that not only passengers can conveniently embark or disembark but also the airline companies can minimize the cost in the whole operational process. Efficient airport operation largely depends on how to gate aircrafts in a smooth flow of arriving and departing flights.

Different kinds of analytical models have been develop-ed on gate assignment problem, such as Mangoubi and Mathaisel 1985, Vanderstraetan 1988, Cheng 1997, Haghani and Chen 1998. At the same time, various techniques have been applied to solve this problem. For instance, linear binary programming (Babic 1984), 0 – 1 linear programming (Bihr 1990), genetic algorithm (Gu & Chung 1999), mixed 0 – 1 quadratic integer programming and tabu search (Xu & Bailey 2001), multi-objective programming (Yan & Huo 2001), simulated annealing (Ding 2002), stochastic programming (Lim & Wang 2005). Most of these techniques are employed to minimize the passenger's walking distance.

The rest of the paper is organized as the follows. First, we formulate the problem employing the techniques of both constraint programming and linear programming with the objective to minimize the conflicts between any two adjacent aircraft assigned to the identical gate. Next, we describe the implementation and experiments with the data of Continental Airlines, Houston IAH airport under specified assumptions. Moreover, we interpret experimental results in detail, which further demonstrate the power and significance of our model. Finally, our conclusion is presented.

## II. PROBLEM FORMULATION

We formulate the airport gate assignment problem as the constraint resource assignment problem where gates serve as the limited resources and aircrafts play the role of resource consumers.

The operation constraints consist of two items: 1) every aircraft must be assigned to one and only one gate. Namely, for a given gate it can be occupied by one and only Chendong Li is with the Department of Computer Science, Texas Tech University, Lubbock, TX 79401 USA (phone: 806-742-3513 ext.236; fax: 806-742-3519; e-mail: chendong.li@ttu.edu).

flight at the same time. Also it can be free without holding any flight at curtain time. 2) For safety consideration, it is prohibited that any two aircrafts are assigned to the same gate simultaneously. In other words, if a gate is occupied by one aircraft, it cannot be assigned to another one until it has been released.

In fact, the airport gate assignment is a very complicated process; while for the sake of simplifying the problem, we mainly take into consideration of the following three factors:
- Number of flights of arriving and departure
- Number of gates available for the coming flight
- The flight arriving and departure time based on the fight schedule

*Modeling*

To model the gate assignment problem in the mathematical form, we first describe the following parameters and data sets.

Gates Set (Resources): $G = \{g_1, g_2, \cdots, g_c\}$ where c is the number of available gates;

Aircrafts Set (Consumers): $F = \{f_1, f_2, \cdots, f_n\}$ where n is the number of aircrafts. For every aircraft $f_i$ ($1 \leq i \leq n$),
- $a_i$: scheduled arriving time
- $d_i$: scheduled departure time
- $b$: buffer time (constant)

Buffer time will lock the gate that has already been assigned to a certain aircraft before it arrives at the gate and after it leaves the gate. The goal of buffer time is to enlarge the interval between any two adjacent aircrafts assigned to the same gate, which will naturally decrease the probability of conflict between these two aircrafts.

We define the *time interval* as the gap of gate locking time between two adjacent aircrafts $f_i$ and $f_j$ and the relation adjacent is defined as any aircrafts $f_i$ and $f_j$ that are assigned to the same gate consecutively. In other words, a certain gate is first occupied by $f_i$ and then sequentially by $f_j$, which indicates that there is no aircrafts assigned to this gate between $f_i$ and $f_j$. According the definition, the time interval locked for a particular aircraft (in terms of a given gate) equals to $[a_i-b, d_i+b]$.

*Decision variable $x_{i,k}$.*

$x_{i,k} = 1$ if and only if aircraft $f_i$ is assigned to gate $c_k$;
$x_{i,k} = 0$ otherwise ($1 \leq i \leq n$, $1 \leq k \leq c$).

*Auxiliary variable $y_{i,j}$.*

$y_{i,j} = 1$ if $\exists k$, $x_{i,k} = x_{j,k} = 1$ ($1 \leq k \leq c$);
$y_{i,j} = 0$ otherwise ($1 \leq i, j \leq n$).

DEFINITION *Conflict*

A conflict or gate conflict is the scenario that it must lead to a collision between any two adjacent aircrafts because of the unreasonable gate assignment or caused by the real departure and arriving time, such as the delay of scheduled time. A gate conflict between any two aircrafts $f_i$ and $f_j$ if both of the following two conditions hold:
- Aircrafts $f_i$ and $f_j$ are assigned to the same gate, that is $y_{i,j} = 1$;
- There is an overlap between the two time intervals of two adjacent aircraft, that is

$[a_i-b, d_i+b] \cap [a_j-b, d_j+b] \neq \varnothing$, which is equivalent to $y_{i,k}*y_{j,k}(d_i-a_j)(d_j-a_i) \leq 0$ ($\forall 1 \leq i, j \leq n$, $i \neq j$, $\forall 1 \leq k \leq c$).

With the objective to minimize the number of gate conflicts which depend on the gate assignment and the scheduled time, we use $p(i,j)$ defined as the probability distribution function on gate conflict between two aircrafts $f_i$ and $f_j$ if they are assigned to the same gate. Specifically, in this model, it equals to $2b/(a_i-d_j+2b)$. Then the gate assignment model can be formulated as follows:

$$Min \ Conflict = \sum_{i,j \in N; i \neq j} \frac{y_{i,j}}{a_i - d_j + 2b} \quad (1)$$

Subject to

$$\sum_{i \in N; k \in C} x_{i,k} = 1 \quad (2)$$

$$\sum_{i,j \in N; i \neq j; k \in C} (x_{i,k} * x_{j,k}) = y_{i,j} \quad (3)$$

$$y_{i,k} * y_{j,k} * (d_i - a_j) * (d_j - a_i) \leq 0 \quad (4)$$

$$x_{i,k} \in \{0,1\} \quad (5)$$

$$\forall 1 \leq i, j \leq n, i \neq j, \forall 1 \leq k \leq c \quad (6)$$

In the model, N is the integer set of the numbers of all the flights needed to be assigned to gates and C stands for the set consisting of all the gates available to host flights. $f_i$ and $f_j$ means different flights and c is the number of gates.

Equation (1) denotes the objective function, which has been simplified by using $1/(a_i-d_j+2b)$ to substitute for

$E(p(i, j))$ in the original objective function[1]. In other word, we applied the following equation to our current objective function (1).

$$E(p(i,j)) = \frac{1}{a_i - d_j + 2b} \quad (7)$$

Equation (2) indicates that each aircraft is assigned to one and only one gate. Equation (3) represents the numerical relationship between $x_{i,k}$ and $y_{i,j}$, which presents a method to compute the auxiliary variable $y_{i,j}$ from $x_{i,k}$. Equation (4) guarantees that one gate can only be assigned to one and only one aircraft at the same time. The scenario that there is an overlap between any two adjacent aircrafts is stated in this constraint. Some additional constraints in the real operations, such as that some particular aircrafts should be assigned to certain gates (like reserved gates and VIP gates), are out of consideration in the current formulated model. Equation (5) represents the decision variables, of which the value is binary.

---

[1] The objective function can be formulated in the form of $\sum_{i,j \in N}(y_{i,j}*E(p(i,j))) = 1$

The proposed model can provide and efficient evaluation criteria for airlines to evaluate the current gate assignment. For example, if an airline authority wants to evaluate the efficiency of the gate assignment of certain number of flights (published as timetable or schedule for passengers' reference) at certain airport, he or she can calculate the value of the objective function in our proposed model based on the published schedule. Intuitively once the value is big, such as bigger than 10, it indicates that the current gate assignment is not good and the authority should consider the reassignment or modify current flight schedule. However, if the value is quite small, such as very near to 0, it denotes that the current gate assignment is almost the desired case in the scenario that the number of available gate is fixed at present.

III. IMPLEMENTATION

Using the Optimization Programming Language we encode our model into OPLscript as shown in Fig.1 and run the program in ILOG OPL studio 3.7.1. In the OPLscript of Figure 1, arrtm, dptm, nbFlt, and nbGate stand for arriving time, departure time, number of Flight and number of Gate, respectively.

We run our program on Dell server PE 1850 under the configuration of Intel(R)Xeon(TM) CPU 3.20GHz, 3.19G Hz, 2.00G of RAM.

```
int nbFlt = ...;
int nbGate = ...;
range Gate 1 .. nbGate;
range Flt 1 .. nbFlt;
var
   int assign[Flt,Gate] in 0..1,
   int y[Flt,Flt] in 0..1;
float+ arrtm[Flt] = ...;
float+ dptm[Flt] = ...;
minimize
     sum (i in Flt, j in Flt : arrtm[j] - dptm[i] >0 ) y[i,j] / (arrtm[j] - dptm[i])
subject to {
        forall(i in Flt, j in Flt : i<>j )
                sum(k in Gate)assign[i,k] * assign[j,k] = y[i,j];
        forall(i in Flt)
                sum(k in  Gate) assign[i,k] = 1;
        forall(i in Flt, j in Flt, k in Gate : i<>j)
                y[i,k]*y[j,k]*(arrtm[i] - dptm[j])*(arrtm[j] - dptm[i]) <= 0;
        };
```

Fig.1 Assignment.mod

IV. EXPERIMENT

In this part we will describe how we conduct all the experiments and report relevant results. Before starting our formal experiment we first obtain the raw data and analyze the data especially due to the large data size. In the following steps, we run the program and collect the testing data. At the end of this part we refer to our future research directions to improve the experiment.

*A. Data Analysis*

We consider an airport with three gates and a schedule of six aircrafts first. To apply the proposed model, we first calculate the matrix of $E(p(i, j))$ (we use the common accepted buffer time as b = 15 for illustration).

In term of large date size, we choose the data of Continental Airlines as our raw data. Based on the timetable of Continental we extract the departure time of all the flights in a whole day, from 6:00A.M. to 23:59P.M., leaving from Houston George Bush International Airport -IAH.

From the schedule, we obtain the departure time of 996 flights in all, in which we did not separate the real timetable that might be different from day to day. For instance, a certain fight may fly to a given destination on special days in a week, like Monday, Wednesday and Friday. In our experiment, we add all the flights with different flight number to out flight set, which will be considered for the gate assignment. Also, we assume that all the flights will leave the airport an hour later after their arrivals, i.e. every flight will stay in the airport for one hour.

The total number of the flight to be considered in the experiment is 996. We plot 996 flights into the coordinate system with arrtm (shorted for arriving time) as vertical axes and x (denoting the number of flights) as horizontal axes, as illustrated in Figure 2. Each small circle in the scatter plot diagram stands for one single flight.

Experiment 1: Using small data set to make sure the model can get the right output.

Experiment 2: Medium data set with changing number of gate to compare the running time and objective value. We assign 33 flights with different number of gates ranging from 1 to 10 and finally 50 as shown in table 1.

Experiment 3: Large data set to compare the running time and objective value. We attempt to assign 996 flights to 70 gates and 65 gates. As the huge search space, it costs even more than two weeks to obtain the final results.

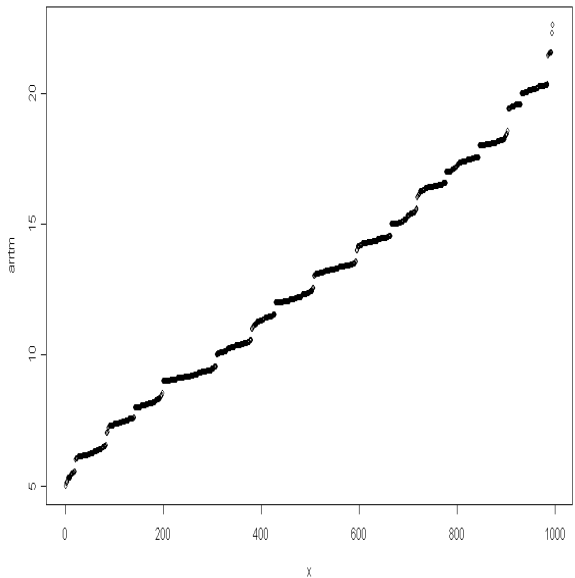

Figure 2 996 Flights scatter plot diagram

## B. Experimental Results

In experiment with small data set, the optimal solution with objective value is 287.0787 indicating that the gate conflicts are inevitable because of the number of available gate is too small. When we enlarge the gate number to 6, the gate conflict decreases dramatically and reaches the value smaller than 3.8615, which is much better compared to 3 gates.

TABLE 1
VARIED GATE ASSIGNMENT

| Number of Gates | Running Time(sec.) | Value of Objective Function |
|---|---|---|
| 1 | 0.03 | 878.0319 |
| 2 | 153.85 | 169.4055 |
| 3 | 311.75 | 28.7304 |
| 4 | 874.01 | 3.5644 |
| 5 | 3.895 | 0 |
| 6 | 4.53 | 0 |
| 7 | 5.19 | 0 |
| 8 | 5.38 | 0 |
| 9 | 5.66 | 0 |
| 10 | 6.17 | 0 |
| 15 | 7.73 | 0 |
| 20 | 9.61 | 0 |
| 30 | 13.86 | 0 |
| 50 | 23.19 | 0 |

Note: The data is based on 33 flights assignment. The Running time presents the average running time.

Table 1 shows the running time and the value of objective function due to the different number of gates available. From this table we can see clearly that when the number of gate equals to 4, the running time is as high as 874.01, the maximum value of the current experiment. When the gate number increases from 5 to 50, the running time augments very slowly. Moreover, it indicates that when the gate number is 5 it is ideal to avoid the gate conflict. In fact, airlines may choose 4 gates considering the large cost of leasing or buying a new gate.

## C. Future Work

Optimizing the objective function with the conditional probability and binomial distribution is a potential direction. Also, we plan to compare the different models and different tools, like LINGO (LINDO), SAT Solver, Excel Solver, and so on. Add some heuristics to shorten the running time of solving large data size problem.

Another direction is that we attempt to build a precise evaluation criteria for the ability of an aircraft-to-gate assignment to handle the uncertainty in the real daily airport operation, i.e. it is a very common phenomenon that aircrafts always arrive late than the original schedule because of some uncontrollable factors like the weather condition; and to search the most robust airport gate assignment or second most robust airport gate assignment (considering the time expense) accurately and effectively.

Limit the number of conflicts in the objective function, like to a constant M (we can set M=0, which means no conflicts or other practical value) and modify the current schedule is also a practical and significant direction.

## V. CONCLUSION

During the airline daily operations, assigning the available gates to the arriving aircrafts based on the fixed schedule is a very important issue. In this paper, we employ the technique of constraint programming and integrate it with linear programming to propose a novel model. The designed experiments demonstrate that our proposed model is of great significance to help airline companies to estimate and even optimize their current flight assignment. Also the experiment illustrates our model is not only simpler, easy to modify, but also pragmatic, feasible and sound.

## ACKNOWLEDGEMENT

I would like to express my gratitude to Dr. Yuanlin Zhang for his valuable discussion and support during the research..

**Chendong Li** is a M.Sc. student in Computer Science at Texas Tech University. His research interest concentrates constraint programming and integer programming.